\documentclass{article}

\usepackage{caption}
\usepackage{microtype}
\usepackage{graphicx}
\usepackage{subfigure}
\usepackage{booktabs} % for professional tables
\usepackage{hyperref}

     \PassOptionsToPackage{numbers, compress}{natbib}
\usepackage[final]{neurips_2021}

\usepackage[utf8]{inputenc} % allow utf-8 input
\usepackage[T1]{fontenc}    % use 8-bit T1 fonts
\usepackage{hyperref}       % hyperlinks
\hypersetup{
  colorlinks=true
}
\usepackage{url}            % simple URL typesetting
\usepackage{booktabs}       % professional-quality tables
\usepackage{amsfonts}       % blackboard math symbols
\usepackage{nicefrac}       % compact symbols for 1/2, etc.
\usepackage{microtype}      % microtypography
\usepackage{xcolor}         % colors

\usepackage{multirow}
\usepackage{graphicx}
\usepackage{pifont}
\usepackage{pifont}
\usepackage[utf8]{inputenc} % allow utf-8 input
\usepackage[T1]{fontenc}    % use 8-bit T1 fonts
\usepackage{url}            % simple URL typesetting
\usepackage{booktabs}       % professional-quality tables
\usepackage{amsfonts}       % blackboard math symbols
\usepackage{nicefrac}       % compact symbols for 1/2, etc.
\usepackage{microtype}      % microtypography
\usepackage{hyperref}
\usepackage{amsfonts}
\usepackage{amssymb,color}%,amsthm}
\usepackage{bbm}
\usepackage{mathrsfs}
\usepackage{amsmath}
\usepackage{amssymb,amsthm}

\usepackage{enumitem}
\setlist[itemize]{noitemsep, topsep=1pt, label=$\bullet$, leftmargin=*}

\usepackage{amsmath,amsfonts,bm}

\def\eqref#1{(\ref{#1})}
\def\1{\bm{1}}

\DeclareMathAlphabet{\mathsfit}{\encodingdefault}{\sfdefault}{m}{sl}
\SetMathAlphabet{\mathsfit}{bold}{\encodingdefault}{\sfdefault}{bx}{n}

\DeclareMathOperator*{\argmin}{arg\,min}

\usepackage{adjustbox}
\usepackage{lipsum}
\usepackage{color}
\usepackage{wrapfig}
\usepackage{booktabs}
\usepackage{multirow,mathtools} 
\usepackage{algorithm,algpseudocode}
\usepackage{threeparttable}
\usepackage[color=gray!20,textsize=scriptsize]{todonotes}

 \algnewcommand{\algorithmicforeach}{\textbf{for each}}
\algdef{SE}[FOR]{ForEach}{EndForEach}[1]
  {\algorithmicforeach\ #1\ \algorithmicdo}% \ForEach{#1}
  {\algorithmicend\ \algorithmicforeach}% \EndForEach

\usepackage{times}

\DeclareMathOperator*{\minimize}{\text{minimize}}

\DeclareMathOperator*{\st}{\text{subject to}}
\DeclareMathAlphabet\mathbfcal{OMS}{cmsy}{b}{n}
\newcommand{\Def}[0]{\mathrel{\mathop:}=}

\def\remark{\addtocounter{remark}{1}\def\@currentlabel{\theremark}%
\emph{Remark~\theremark}. } \makeatother

\newcounter{remark}

\def\b1{{\boldsymbol 1}}

\def\bx{\mathbf x}

\def\bm{\mathbf m}

\def\by{\mathbf y}

\usepackage{multirow}
\newcommand{\din}{\mathcal D^{\texttt{tr}}}

\newcommand{\dout}{\mathcal D^{\texttt{val}}}

\def\minimize{\mathop{\text{minimize}}}

\usepackage{color, colortbl}
\definecolor{Gray}{gray}{0.9}
\definecolor{Orange}{rgb}{1,0.5,0}
\makeatletter
\newcommand*{\rom}[1]{\expandafter\@slowromancap\romannumeral #1@}
\makeatother
\newcommand{\mycomment}[1]{}

\newcommand{\SignMAML}{{Sign-MAML}}
\newcommand{\FOMAML}{{FO-MAML}}
\newcommand{\MAML}{{MAML}}
\newcommand{\FSCIFAR}{{FS-CIFAR100}}

\title{Sign-MAML:  Efficient Model-Agnostic Meta-Learning by SignSGD}

\author{%
  Chen Fan \\
  College of Information and Computer Sciences, \\
  University of Massachusetts Amherst
  \And
  Parikshit Ram\\
  IBM Research\\
  \And
  Sijia Liu\\
  Computer Science and Engineering, Michigan State University\\
  MIT-IBM Watson AI Lab
}

\begin{document}

\maketitle

\begin{abstract}
%\SL{[Revised]}
We propose a new  computationally-efficient first-order
algorithm  for Model-Agnostic Meta-Learning ({\MAML}). 
The key enabling technique is to interpret {\MAML} as a bilevel optimization (BLO) problem and leverage the sign-based SGD (signSGD)   as a lower-level optimizer of BLO. 
We show that {\MAML}, through the lens of signSGD-oriented BLO, naturally yields an alternating optimization scheme that just requires first-order gradients of a learned meta-model. 
We term the resulting {\MAML} algorithm \textit{\SignMAML}. 
Compared to the conventional first-order  MAML ({\FOMAML}) algorithm, {\SignMAML} is theoretically-grounded as it does not impose any  assumption on the absence of second-order derivatives during meta training.
In practice, we show that {\SignMAML} outperforms {\FOMAML} in various few-shot image classification tasks, and  compared to {\MAML}, it achieves a much more graceful tradeoff between classification accuracy and computation efficiency. 

% This is in contrast to MAML and 
% We apply this finding to MAML and show that lower-level signSGD unrolling leads to the same meta-gradient as that of first-order MAML (FO-MAML). We name MAML with signSGD unrolling as Sign-MAML, and highlight that the difference between Sign-MAML and FO-MAML is not only in the lower-level optimizer, i.e. signSGD for Sign-MAML and SGD for FO-MAML, but also in their treatments of second-order derivatives. Sign-MAML naturally does not require second-order derivatives, whereas FO-MAML explicitly neglects them in the computation. We empirically study the performance of Sign-MAML and show that it can out-perform FO-MAML and MAML at high computation efficiency. 

\end{abstract}

\section{Introduction}
% \SL{[Revised] Please use signSGD instead of SignSGD.}

Humans can learn new tasks quickly based on prior knowledge or experience with similar tasks. A meta-learning algorithm resembles this in a way such that given previous exposure to relevant tasks, new tasks can be learned with a small amount of data. To do this, it involves a meta(or upper)-learner whose job is to update parameters of a base(or lower)-learner which aims to solve a specific task (e.g. image classification) at hand. This `learning to learn' hierarchical structure can be viewed as solving a bilevel optimization (BLO) problem, in which the solution to the lower-level problem provides useful feedback for updating the solution of an upper-level problem \citep{vanschoren2018meta, hospedales2020meta}. Recent works have studied the optimization-based meta-learning approach targeting on different parameters associated with the base learner, such as learning a good weight initialization \citep{finn2017model, nichol2018firstorder, raghu2019rapid}, and updating neural network architectures  \citep{zoph2016neural, elsken2020meta, lian2019towards}.
%or proposing step rules in case of the lower-level optimizer being gradient descent (GD) \citep{ravi2016optimization, andrychowicz2016learning}.  

Within the optimization-based meta-learning family, Model-Agnostic Meta-Learning (MAML) is a popular method that has been widely applied to solving computer vision and natural language processing  tasks \citep{zintgraf2019fast, dou2019investigating, liu2020does}. 
In-depth empirical and theoretical understanding of MAML has also been provided in 
%Its empirical and convergence properties in nonconvex settings have been investigated in several works 
\citep{nichol2018firstorder,antoniou2018train, fallah2020convergence, ji2020convergence}.
Through the lens of BLO,
{\MAML} is composed of an upper-level optimization step (which updates weight initialization of a model), and a sequence of lower-level steps (which adapt this initialization to different specific tasks).
Despite the effectiveness of {\MAML}, it is difficult to scale to large models and datasets due to the need of  
second-order derivatives during model training \citep{nichol2018firstorder, rajeswaran2019meta}. A first-order variant (FO-MAML) solves this problem by ignoring \citep{finn2017model} or estimating \citep{fallah2020convergence} second-order derivatives in practice at the cost of introducing meta-gradient estimation error. 
%\todo{Mention that \citep{fallah2020convergence} also propose HF-MAML that also doesn't need Hessian but does better than FO-MAML}

% We propose to use signSGD as the lower-level optimizer of MAML (Sign-MAML) and show that this is indeed the case. 
% We unify the understand of these methods and compare them in the framework of bilevel-optimization.

\paragraph{Contributions} 
In this work, we aim to design a computationally-efficient and 
theoretically-grounded   MAML algorithm 
that only relies on first-order derivatives in its implementation.  To this end, we propose {\SignMAML} by integrating MAML with signSGD \citep{bernstein2018signsgd} and show its advantages over {\FOMAML} and {\MAML}.
Our contributions  are summarized below:

\begin{itemize}
\item (Formulation-wise) In \S \ref{sec:formulation}, we revisit {\MAML} through the lens of BLO and identify a tight connection between its computation efficiency and the choice of a lower-level optimizer.

\item (Methodology-wise) 
In \S \ref{sec:method}, we leverage signSGD to unroll the lower-level problem of {\MAML} and theoretically show that this naturally leads to a first-order alternating optimization method, {\em Sign-MAML}, whose computation is exactly as efficient as {\FOMAML}. 
%Applying this result to MAML, we show that its upper-level update no longer requires computing second-order derivatives hence alleviate its computation and memory burdens. 

%$\bullet$ We theoretically show that Sign-MAML and FO-MAML enjoy the same computation efficiency, 
%but the former 
%using different lower-level optimizers. We remark that Sign-MAML naturally obtains this result whereas FO-MAML has to make additional assumptions. 

\item (Application-wise) In \S \ref{sec:application}, we conduct extensive experiments to demonstrate the advantage of Sign-MAML in computation efficiency and accuracy. 
% In particular, it takes a similar computation cost as FO-MAML but outperforms FO-MAML in accuracy for challenging few-shot learning tasks. 
In particular, we show that Sign-MAML has computational costs similar to those of FO-MAML while providing significantly improved accuracy for few-shot tasks.
%One the other hand,   it can obtain comparable or even better accuracy than  MAML but is   computationally-light. 
\end{itemize}

\section{Related Work}
\paragraph{Meta-learning} 
A surge of recent works have been devoted to developing
theory and algorithms of {\MAML} \citep{hospedales2020meta,raghu2019rapid,fallah2020convergence,ji2020convergence,song2019maml}.  For example, 
the `Almost No Inner Loop' (ANIL) algorithm was proposed in \citep{raghu2019rapid}, which dissects the meta-learning into two phases: training the  initialization of a meta-model, and partially fine-tuning the classification head of the meta-model.  Compared to the conventional {\MAML} algorithm, ANIL yields  a reduced computation cost due to the use of partial fine-tuning instead of the end-to-end full fine-tuning. However, ANIL still needs second-order derivatives during meta training. To overcome such a computation bottleneck, the work \citep{fallah2020convergence,song2019maml} proposed to use the finite difference of function values or first-order gradients to estimate the high-order derivatives involved in {\MAML}. However, the resulting gradient/Hessian estimation may not be unbiased  and could lead to an unexpected large estimation variance \citep{liu2020primer}.  
% In a recent work, Raghu et al. investigated how much adaptation actually happens during the lower-level updates of MAML, and they found that the meta-initialization learned at upper-level already contains many useful feature representations. Hence this meta-initialization can be quickly adapted to specific tasks without significant changes. They  proposed "Almost No Inner Loop" (ANIL) algorithm which only updates the head of a neural network at both lower-level and upper-level \citep{raghu2019rapid}. The ANIL algorithm is more efficient than MAML as a big fraction of weights (i.e. a neural network body) is not updated in lower-level updates. However, its performance may not be as good as MAML. 
% Ji et al. provided an analysis on this algorithm for both strongly convex and nonconvex lower-level objectives \citep{ji2020convergence}.
Another first-order method proposed in \citep{nichol2018firstorder} simplifies meta gradient computation by using the difference between initial and adapted weights. Here, we focus on the design of lower-level optimizer to speed up computation. Besides optimization-based meta-learning, other algorithms such as metric-based and model-based meta-learning have also been developed \citep{hospedales2020meta,koch2015siamese, snell2017prototypical, sung2018learning, vinyals2016matching,santoro2016meta}.  In this work, we focus on the optimization-based meta-learning. 
% The metric-based algorithms involve comparing data points in the training and validation set to make predictions \citep{hospedales2020meta}. Examples are matching, siamese, prototypical and relation neural networks \citep{koch2015siamese, snell2017prototypical, sung2018learning, vinyals2016matching}. Model-based algorithms involve using a model to encode information of a dataset \citep{hospedales2020meta}. One example is memory-augmented neural networks \citep{santoro2016meta}.  
%
\paragraph{Bilevel optimization}
Bilevel optimization is applied to solve problems that exhibit two-level hierarchical structure in which the solution to the lower-level problem is an input to the upper-level problem. Solvers for BLO problems can be either deterministic or stochastic. Under deterministic BLO appraoch, two commonly used methods are approximate implicit differentiation (AID) based and iterative differentiation (ITD) based. For both methods, the lower-level problem is solved by gradient descent (GD). For the upper-level problem, AID-based methods obtain meta-gradients through implicit gradients~\citep{gould2016differentiating} whereas ITD-based methods rely on backpropagation~\citep{grazzi2020iteration, ji2021bilevel}. In recent years, stochastic approaches have gained a lot of attentions due to its fast convergence and scalability. \citet{ghadimi2018approximation} proposed a method to obtain lower and upper-level gradients through stochastic approximations. \citet{hong2020two} developed a method that solves lower and upper-level problems simultaneously  with lower and upper-level step sizes at two different scales.  

\paragraph{Sign-based optimization methods}
%signSGD was proposed in   \citep{bernstein2018signsgd} for the first time, by taking
signSGD~\citep{bernstein2018signsgd} utilizes the sign of gradients as the descent direction for FO non-convex optimization and demonstrates
%\citep{bernstein2018signsgd} showed that signSGD    can achieve 
a convergence rate comparable to that of stochastic gradient descent (SGD). \citet{liu2018signsgd} proposed zeroth-order signSGD (ZO-signSGD) for solving optimization problems where first-order derivatives are difficult or infeasible to obtain, demonstrating lower estimation variance when compared to conventional ZO-SGD schemes \citep{liu2020primer,ghadimi2013stochastic}. %\citep{ghadimi2018approximation}.
% Compared to the conventional ZO-SGD algorithm, it was shown that taking the sign of a rough gradient estimate helps reduce the estimation variance.
In adversarial machine learning, fast gradient sign method (FGSM)   has   been commonly used for generating prediction-evasion adversarial attacks  \citep{goodfellow2014explaining} and for training an adversarially robust deep neural network \citep{madry2017towards}.

%\todo{This line ends funny}
% Liu et al. further developed a zeroth-order signSGD (ZO-signSGD) algorithm and applied it  to generate adversarial samples from a deep neural network \citep{liu2018signsgd}. 
% FGSM. 

\section{Problem Statement} \label{sec:formulation}
In this section, we begin by presenting the problem of MAML and interpreting it through the lens of BLO (bilevel optimization). Next, we illustrate the limitations of existing   solutions to MAML and elaborate on our research objective.

\paragraph{BLO setup of MAML}
 Considering $P$ tasks where each task $\tau_i, i \in [P]$ is sampled from a task distribution, % $p_{\tau}$,
MAML seeks to solve the following problem from a bilevel optimization perspective:
\begin{align}
\begin{array}{ll}
\displaystyle \minimize_{\mathbf x}     & L(\mathbf x) \Def  
%\frac{1}{N} \sum_{i=1}^N \ell_i(\mathbf x)= 
\frac{1}{P} \sum_{i=1}^P \ell_i(\mathbf{y}_i^*(\mathbf x) ; \dout_i)  \\
   \st   & \mathbf{y}_i^*(\mathbf x) \in
 \displaystyle  \argmin_{ \mathbf y^\prime  } \, \ell_i(\mathbf{y}^\prime; \din_i ), \quad \forall i \in [P] = \{1, \ldots, P\},
\end{array}
%\tag{{\MAML}}
\label{eq: MAML_obj}
\end{align}
where $\mathbf x$ is the weight initialization of a model, $\mathbf{y}_i^*(\mathbf x)$ is the optimal weight after the model is fine-tuned for task $i$ with $\ell_i$, $\din_i$ and $\dout_i$ as the task-specific loss, training (support) set and validation (query) set respectively.
% $\ell_i$ is the loss of task $i$, and $\din_i$ and $\dout_i$ are task-specific training (support) or validation (query) datasets, respectively. 

A generic BLO formulation of the MAML problem \eqref{eq: MAML_obj} is then given by
%has the following form: 
%\begin{align}
%    \begin{array}{ll}
%\displaystyle \minimize_{\mathbf x\in \mathbb R^{d_1}}         & \underbrace{f(\mathbf x, %\mathbf y^*(\mathbf x))}_\text{Upper-level problem} \vspace*{0.5mm}\\
%  \st & \underbrace{ \displaystyle \mathbf y^*(\mathbf x) = \argmin_{\mathbf y}  g(\mathbf x, %\mathbf y)}_\text{Lower-level problem},
%    \end{array}
%    \label{eq: prob_bi_level}
%\end{align}
\begin{align}
\displaystyle \minimize_{\mathbf x}
\underbrace{f(\mathbf x, \mathbf y^*(\mathbf x))}_\text{Upper-level problem} 
\st 
\underbrace{ \displaystyle \mathbf y^*(\mathbf x) \in \argmin_{\mathbf y}  g(\mathbf x, \mathbf y)}_\text{Lower-level problem},
\label{eq: prob_bi_level}
\end{align}
where   a lower-level solution    is used as an input to minimize the upper-level objective $f$.
The MAML problem \eqref{eq: MAML_obj} is a special case of the BLO formulation  \eqref{eq: prob_bi_level}:
The upper-level objective function $L(\mathbf x)$ in \eqref{eq: MAML_obj} is not an exact bi-variate function $f(\mathbf x, \mathbf y)$ as \eqref{eq: prob_bi_level}; Instead, $L(\mathbf x)$ relies only on a  lower-level solution $\mathbf y^*$, which is a function of $\mathbf x$. 

\paragraph{Second-order derivatives requested in {\MAML}}
Conventionally, MAML \citep{finn2017model} solves 
 the lower-level problem of  
\eqref{eq: MAML_obj}  through a $m$-step \textbf{SGD unrolling}. Let  $\ell_i(\mathbf{y}^\prime ) \Def  \ell_i(\mathbf{y'} ; \dout_i) $, $\hat{\ell}_i(\mathbf{y}^\prime ) \Def \ell_i(\mathbf{y}^\prime;   \din_i)$, and 
$\mathbf x_k$ denote the model initialization at the $k$th upper-level iteration, the  original MAML algorithm is then given by
\begin{align}
  &  \text{Lower-level: }
  \underbrace{ \mathbf{y}_i^{(0)}(\mathbf x_k) = \mathbf x_k; \hspace{0.5 em} \mathbf{y}_i^{(m)}(\mathbf x_k) = \mathbf{y}_i^{(m-1)}(\mathbf x_k) - \beta \left. \nabla_{\mathbf y^\prime} \hat{\ell}_i(\mathbf y^\prime) \right |_{\mathbf y_i^{(m-1)}(\mathbf x_k)}}_\text{$m$-step SGD unrolling}
    \label{eq: sgdInner}\\
  &  \text{Upper-level: } \mathbf x_{k+1} = \mathbf x_{k} - \alpha \frac{1}{P} \sum_{i=1}^P  \nabla_{\mathbf x} \ell_i(\mathbf y_i^{(m)}(\mathbf x_k)),
\end{align}
where $\alpha, \beta > 0$ are learning rates of SGD used for upper-level and lower-level optimization, respectively. 
Substituting the lower-level SGD unrolling into the upper-level SGD   step, the overall optimization step to update the optimizee variable $\mathbf x$
is given by \citep{finn2017model, nichol2018firstorder}
\begin{align}
    \mathbf{x}_{k+1} = \mathbf{x}_k - \alpha \underbrace{  \frac{1}{P} \sum_{i=1}^P \prod_{{{n=0}}}^{{m-1}} \left( \mathbf I - \beta \left. \nabla^2_{\mathbf y^\prime} \hat{\ell}_i(\mathbf y^\prime) \right |_{\mathbf y^\prime = \mathbf{y}^{(n)}_i(\mathbf x_k)} \right) \left. \nabla_{\mathbf y^\prime} \ell_i(\mathbf y^\prime) \right |_{\by' = \mathbf{y}_i^{(m)}(\mathbf x_k)} }_\text{Meta-gradient w.r.t. $\mathbf x$},
    \label{eq: backMAML}
\end{align}
where $\nabla_{\mathbf y^\prime}^2$ denotes the second-order derivatives with respect to (w.r.t.) the variable $\mathbf y^\prime$.
In \eqref{eq: backMAML},
the computation involving $\nabla^2_{\mathbf y^\prime} \hat{\ell}_i(\mathbf y^\prime)$ is \textbf{costly} for large neural networks and datasets, and this cost increases with the number of fine-tuning steps.  

\paragraph{Research objective}
To resolve the difficulty induced by second-order derivatives, FO-MAML \textbf{assumes} them to be $\mathbf 0$ in the computation \citep{finn2017model}. This would introduce an error into the meta-gradient in \eqref{eq: backMAML}, and may hamper its   generalization ability. In fact, a generalization gap as large as $6 \%$ is observed for tasks such as 20-way 1-shot on Omniglot dataset between MAML and FO-MAML   \citep{rajeswaran2019meta}.
\citet{fallah2020convergence} present a Hessian-free MAML which has improved theoretical convergence over FO-MAML but its empirical performance or efficiency has not been studied.
From (\ref{eq: backMAML}), we see the cause of high computation cost is rooted in the coupling between lower and upper-level problems in which backpropagation has to loop through the entire lower-level optimization trajectory.
% Besides {\MAML},
To bypass this, \citet{rajeswaran2019meta} proposed implicit MAML (iMAML) to directly solve the BLO problem \eqref{eq: prob_bi_level} using the implicit gradient method. 
%decouples these two problems by introducing a regularized lower-level objective and computing meta-gradient through implicit gradients \SL{[citation]}. 
However,
%\textcolor{blue}{even more computationally-intensive than {\MAML}}\todo{maybe remove this part of the line since this depends on the number of inner-level GD steps},
iMAML    needs (an approximation of) a matrix inversion operation to calculate an implicit gradient.
% are done with numerical algorithms such as conjugate gradient (CG). The iMAML approach still requires computing Hessian vector products and additional hyperparameter tuning for the regularization parameter as well as the number of CG steps \citep{rajeswaran2019meta}.
Among the aforementioned   algorithms, {\FOMAML} is the computationally lightest but yields a poorer optimization accuracy. By contrast, {\MAML} and {iMAML} have improved generalization ability but higher computational costs. 
Spurred by above, we ask:

\begin{center}
    \textit{How to develop an assumption-least first-order {\MAML} algorithm that enjoys the dual advantages of low computation cost and high optimization accuracy?}
\end{center}

% Therefore, our research goal is to design a new first-order approach to gain computation-efficiency  without incurring performance loss.

\section{Sign-MAML: Advancing MAML by SignSGD} \label{sec:method}
In this section, we first 
present the  method of signSGD unrolling to solve the 
 BLO problem \eqref{eq: prob_bi_level}. Then we apply the achieved results to the case of MAML to   establish our Sign-MAML method. At the end, we highlight the differences between FO-MAML and Sign-MAML.

\paragraph{BLO solver based on signSGD unrolling}
 We propose to unroll the lower-level problem in \eqref{eq: prob_bi_level} using signSGD \citep{bernstein2018signsgd}. The last step of a $m$-step unrolling via signSGD is given by 
\begin{align}
     \mathbf y^{(m)} =  \mathbf y^{(m-1)} - \beta \mathrm{sign}(\nabla_{\mathbf y} g(\mathbf x, \mathbf y^{(m-1)})),
     \label{eq: sign_SGD_unrolling}
\end{align}
where $\mathrm{sign}(\cdot)$ denotes element-wise sign operation, $\beta > 0$ is the lower-level learning rate, and $\by^{(0)}$ can be a random starting point. 
Substituting \eqref{eq: sign_SGD_unrolling} into problem \eqref{eq: prob_bi_level} with $\mathbf y^*(\mathbf x) = \mathbf y^{(m)}$, we have the following variant of the original BLO problem
\begin{align}
    \begin{array}{ll}
\displaystyle \minimize         & f(\mathbf x, \mathbf y^{(m)}(\mathbf x)) ,
    \end{array}
    \label{eq: prob_bi_level_unroll}
\end{align}
%\SL{[Why needs $d_1$?]}
where we explicitly express $\mathbf y^{(m)}$ as a function of $\mathbf x$. To optimize $\mathbf x$, we resort to GD/SGD
\begin{align}
    \mathbf x_{k+1} = \mathbf x_{k} - \alpha
    \frac{d f(\mathbf x_{k}, \mathbf y^{(m)}(\mathbf x_{k}))}{ d \mathbf x},
   % \nabla_{\mathbf x} f(\mathbf x, \mathbf y^{(m)}(\mathbf x)),
\label{eq: GD_signMAML}
\end{align}
where $\alpha > 0$ is the upper-level learning rate and $k$ is the descent step index. In \eqref{eq: GD_signMAML},
the key step is to compute the gradient w.r.t. $\mathbf x$, namely,
\begin{align}
    \frac{d f(\mathbf x_{k}, \mathbf y^{(m)}(\mathbf x_{k}))}{ d \mathbf x} = \frac{\partial f(\mathbf x_{k}, \mathbf y^{(m)}(\mathbf x_{k}))}{\partial \mathbf x} + \frac{d \mathbf y^{(m)}(\mathbf x_{k})^\top}{d \mathbf x} \frac{\partial f(\mathbf x_{k}, \mathbf y^{(m)}(\mathbf x_{k}))}{\partial \mathbf y}, \label{eq: grad_unroll}
\end{align}
where $\frac{\partial f (\bx, \by)}{\partial \bx} = \nabla_{\bx} f$ 
and $\frac{d f (\bx, \by)}{d \bx}$ denote the partial and full derivatives of $f$ w.r.t. the variable $\bx$ respectively. 
Based on 
 \eqref{eq: sign_SGD_unrolling} and the \textbf{key fact} that 
 $ \frac{d \mathrm{sign}(\mathbf z)^\top}{d \mathbf z} = \mathbf 0$ (holding almost surely), we have
 %that $ \mathbf y^{(m)} (\mathbf x_{k})=  \mathbf y^{(m-1)}(\mathbf x_{k}) - \beta \mathrm{sign}(\nabla_{\mathbf y} g(\mathbf x_{k}, \mathbf y^{(m-1)}(\mathbf x_{k})))$, we thus have 
\begin{align}
%\label{eq: chain_rule_sign}
     \frac{d \mathbf y^{(m)}(\mathbf x_{k})^\top}{d \mathbf x} 
    %  & = 
    %   \frac{d \mathbf y^{(m-1)}(\mathbf x_{k})^\top}{d \mathbf x} - \beta 
    %   \frac{d \nabla_{\mathbf y} g(\mathbf x_{k}, \mathbf y^{(m-1)}(\mathbf x_{k}))^\top}{d \bx}
    %   \frac{d \mathrm{sign}(\mathbf z)^\top}{d \mathbf z}  \Bigr|_{\mathbf z = \nabla_{\mathbf y} g(\mathbf x_{k}, \mathbf y^{(m-1)}(\mathbf x_{k}))} \nonumber \\
      & =  \frac{d \mathbf y^{(m-1)}(\mathbf x_{k})^\top}{d \mathbf x} = \cdots = \frac{d \mathbf y^{(0)}(\mathbf x_{k})^\top}{d \mathbf x}.
          \label{eq: unroll_grad}
\end{align}
% where we have used the fact that $ \frac{d \mathrm{sign}(\mathbf z)^\top}{d \mathbf z} = \mathbf 0$ is valid at almost everywhere except for $\mathbf z = \mathbf 0$. Based on this, we eventually have
% \begin{align}
%     \frac{d \mathbf y^{(m)}(\mathbf x_{k})^\top}{d \mathbf x} = \frac{d \mathbf y^{(0)}(\mathbf x_{k})^\top}{d \mathbf x}.
%     \label{eq: unroll_grad}
% \end{align}

Substituting \eqref{eq: unroll_grad} into
\eqref{eq: grad_unroll}, we achieve
\begin{align}
     \frac{d f(\mathbf x_{k}, \mathbf y^{(m)}(\mathbf x_{k}))}{ d \mathbf x} = \frac{\partial f(\mathbf x_{k}, \mathbf y^{(m)}(\mathbf x_{k}))}{\partial \mathbf x} +  \frac{d \mathbf y^{(0)}(\mathbf x_{k})^\top}{d \mathbf x} \frac{\partial f(\mathbf x_{k}, \mathbf y^{(m)}(\mathbf x_{k}))}{\partial \mathbf y}.
\end{align}
This implies that when signSGD is used to unroll the lower-level problem, we can naturally reach a \textbf{first-order alternating optimization} method:
\begin{align}
&    \text{$\by$-step: signSGD unrolling \eqref{eq: sign_SGD_unrolling}} \label{eq: y_step_unroll}\\
&  \text{$\bx$-step: } 
  \mathbf x_{k+1} = \mathbf x_{k} - \alpha
    \frac{\partial f(\mathbf x_{k}, \mathbf y^{(m)}(\mathbf x_{k}))}{ \partial \mathbf x} - \alpha \frac{d \mathbf y^{(0)}(\mathbf x_{k})^\top}{d \mathbf x} \frac{\partial f(\mathbf x_{k}, \mathbf y^{(m)}(\mathbf x_{k}))}{\partial \mathbf y}.
    \label{eq: x_step_unroll}
\end{align}

% \begin{wrapfigure}{R}{0.5\textwidth}
% %\begin{figure}[H]
% \centering
% \vspace*{-4mm}
%     \begin{minipage}{0.5\textwidth}
%       \begin{algorithm}[H]
%           \caption{Sign-MAML}
%         \begin{algorithmic}[1]
%           \For{$k = 1, 2, \ldots$}
%              \State Sample $P$ tasks from a task distribution 
%              \For{$i = 1,2, \ldots, P$}
%              \State Initialize $\mathbf{y}_i^{(0)}(\mathbf x_k) = \mathbf x_k$
%              \State Obtain $\mathbf y_i^{(m)}$ by   signSGD unrolling   \eqref{eq: sign_SGD_unrolling} 
%             %  \State $\vdots$
%             %  \State \( \mathbf{y}_i^{(m)}(\mathbf x_k) = \mathbf{y}_i^{(m-1)}(\mathbf x_k) - \beta \left. \mathrm{sign} \nabla_{\mathbf y^\prime} \hat{\ell}_i(\mathbf y^\prime) \right |_{\mathbf y_i^{(m-1)}(\mathbf x_k)} \); 
%          \EndFor
%          \State Compute $\mathbf x_{k+1}$ using \eqref{eq: FO-MAML}; 
%         %\State $k \gets k+1$
%         \EndFor
%     %\EndWhile
%   \end{algorithmic}
%   \label{alg: MAML-SIGN}
%       \end{algorithm}
%     \end{minipage}
%       \vspace*{-5mm}
% \end{wrapfigure}
%\end{figure}
\paragraph{{\SignMAML}: MAML based on signSGD unrolling} 
We now apply (\ref{eq: y_step_unroll}) and (\ref{eq: x_step_unroll}) to the case of MAML in which $\mathbf y^{(0)}(\mathbf x_{k}) = \mathbf x_{k}$ and the dependence of upper objective of MAML on $\mathbf{x}$ is only through $\mathbf y(\mathbf x)$. They lead to $\frac{d \mathbf y^{(0)}(\mathbf x_{k})^\top}{d \mathbf x}  = \mathbf I$ and 
$\frac{\partial f(\mathbf x_{k}, \mathbf y^{(m)}(\mathbf x_{k}))}{\partial \mathbf x} = \mathbf 0$. 
These two simplifications render \eqref{eq: x_step_unroll} to 
%$\mathbf x_{k+1} = \mathbf x_{k} - \alpha  \frac{\partial f(\mathbf x_{k}, \mathbf y^{(m)}(\mathbf x_{k}))}{\partial \mathbf y}$, which can be rewritten as 
\begin{align}
    \mathbf{x}_{k+1} = \mathbf{x}_k - \alpha \frac{1}{P} \sum_{i=1}^P \left. \nabla _{\mathbf y^\prime} \ell_i(\mathbf y^\prime) \right |_{\mathbf y^\prime = \mathbf{y}_i^{(m)}(\mathbf x_k)}.
    \label{eq: FO-MAML}
\end{align}

At the first glance, the upper-level MAML update with signSGD unrolling is the same as {\FOMAML}. However, the \textbf{key difference} lies in the \textbf{choice of lower-level optimizer}:
signSGD unrolling \textbf{naturally} leads to \eqref{eq: FO-MAML}. By contrast, FO-MAML requires making the \textbf{assumption} of $\nabla^2_{\mathbf y^\prime} \hat{\ell}_i(\mathbf y^\prime) = \mathbf{0}$ in \eqref{eq: backMAML}. Hence, we regard using signSGD unrolling as the `authentic' first-order method, and we name our algorithm as {\SignMAML} shown in Algorithm\,1.
We also remark that in addition to signSGD, the gradient sign-based momentum method \cite{bernstein2018signsgd} is another possible alternative to generate the first-order MAML approach via gradient unrolling. 

% \begin{wrapfigure}{R}{0.5\textwidth}
% %\begin{figure}[H]
% \centering
% \vspace*{-4mm}
%     \begin{minipage}{0.5\textwidth}
%       \begin{algorithm}[H]
%           \caption{Sign-MAML}
%         \begin{algorithmic}[1]
%           \While {not done}
%              \State Sample $N$ tasks from a task distribution 
%              \For{$i = 1,2, \ldots, N$}
%              \State Initialize $\mathbf{y}_i^{(0)}(\mathbf x_k) = \mathbf x_k$
%              \State Obtain $\mathbf y_i^{(m)}$ by   signSGD unrolling   \eqref{eq: sign_SGD_unrolling} 
%             %  \State $\vdots$
%             %  \State \( \mathbf{y}_i^{(m)}(\mathbf x_k) = \mathbf{y}_i^{(m-1)}(\mathbf x_k) - \beta \left. \mathrm{sign} \nabla_{\mathbf y^\prime} \hat{\ell}_i(\mathbf y^\prime) \right |_{\mathbf y_i^{(m-1)}(\mathbf x_k)} \); 
%          \EndFor
%          \State Compute $\mathbf x_{k+1}$ using \eqref{eq: FO-MAML}; 
%         \State $k \gets k+1$
%     \EndWhile
%   \end{algorithmic}
%   \label{alg: MAML-SIGN}
%       \end{algorithm}
%     \end{minipage}
% \end{wrapfigure}
% %\end{figure}

% \begin{wrapfigure}{R}{0.5\textwidth}
% %\begin{figure}[H]
% \centering
% \vspace*{-4mm}
%     \begin{minipage}{0.5\textwidth}
      \begin{algorithm}[H]
           \caption{Sign-MAML}
        \begin{algorithmic}[1]
          \For{$k = 1, 2, \ldots$}
             \State Sample $P$ tasks from a task distribution 
             \For{$i = 1,2, \ldots, P$}
             \State Initialize $\mathbf{y}_i^{(0)}(\mathbf x_k) = \mathbf x_k$
             \State Obtain $\mathbf y_i^{(m)}$ by   signSGD unrolling   \eqref{eq: sign_SGD_unrolling} 
            %  \State $\vdots$
            %  \State \( \mathbf{y}_i^{(m)}(\mathbf x_k) = \mathbf{y}_i^{(m-1)}(\mathbf x_k) - \beta \left. \mathrm{sign} \nabla_{\mathbf y^\prime} \hat{\ell}_i(\mathbf y^\prime) \right |_{\mathbf y_i^{(m-1)}(\mathbf x_k)} \); 
         \EndFor
         \State Compute $\mathbf x_{k+1}$ using \eqref{eq: FO-MAML}; 
        %\State $k \gets k+1$
        \EndFor
    %\EndWhile
  \end{algorithmic}
  \label{alg: MAML-SIGN}
      \end{algorithm}
%     \end{minipage}
%       \vspace*{-5mm}
% \end{wrapfigure}

\section{Experiments} \label{sec:application}
The central questions that we aim to address with our experiments are: \textit{\ding{172} Can Sign-MAML perform better than FO-MAML without increasing computation cost? \ding{173} Can Sign-MAML perform comparably to MAML but with less computation time?} To this end, we measure  the \textit{test accuracy} and \textit{train time per upper-iteration}
of Sign-MAML, together with   baselines (FO-MAML and MAML)  in  $N$-way and $K$-shot image classification tasks.
%We compare the performance of Sign-MAML against those of FO-MAML and MAML for different N-way and K-shot tasks. 

\subsection{Experiment setup}
\paragraph{Datesets} We conduct experiments on Fewshot-CIFAR100 ({\FSCIFAR}) and MiniImageNet   datasets \citep{vinyals2016matching, oreshkin2019tadam}.  The {\FSCIFAR} dataset has  600 images of size $32 \times 32$ in each of the $100$ classes from CIFAR100 \citep{krizhevsky2009learning}. We partition the 100 classes into $60$ classes , $20$ classes and $20$ classes for train, validation and test respectively following \citet{oreshkin2019tadam}. 
The MiniImageNet dataset has 600 images of size $84 \times 84$ in each of the $100$ classes. We partition the $100$ classes into $64$ classes, $16$ classes and $20$ classes for train, validation and test respectively following \citet{ravi2016optimization}.

\paragraph{Architectures} For MiniImageNet, we use a neural network consisting of $4$ convolutional layers with $32$ filters in each layer used in \citet{ravi2016optimization}. For {\FSCIFAR}, 
we also use the $4$-layer convolutional neural newtork but with $64$ filters in each layer. For both neural networks, each convolution operation is followed by batch normalization, ReLU activation and $2 \times 2$ max pooling. 

\paragraph{Implementation details} We use $\alpha = 0.001$ as the upper-level learning rate, $m = 1$ fine-tuning step for training (unless otherwise specified), $m = 10$ fine-tuning steps for testing, and $P=4$ as the  batch size  of tasks across all experiments. To setup the lower-level learning rate $\beta$, since different optimizers (e.g., signSGD vs. SGD) are used, 
we perform a grid search on $\beta$ and pick the one with the best validation performance (see Appendix A for details). We utilized the existing implementation of FO-MAML and MAML in the \href{http://learn2learn.net/}{\texttt{learn2learn} Python package}~\citep{arnold2020learn2learn} and adapt it to implement Sign-MAML.\footnote{Our codes are available at \url{https://github.com/chenfan95/Sign-MAML}}
% The code for implementing Sign-MAML was adapted from \citet{arnold2020learn2learn}. 
% We run all FO-MAML and MAML experiments using the code directly from \citet{arnold2020learn2learn}. 

%(see Appendix for details), and present results of those that have the best test performance.
% meta-batch size ($N$) $4$, 1 fine-tuning step for training,

%\begin{wraptable}{r}{80mm}
\begin{table}[htb]
 %\captionsetup{width=11.5cm}
  \caption{
  \footnotesize{{\FSCIFAR} classification results, which include accuracy (upper-level numbers) and computation time per meta-iteration in seconds (lower-level numbers). For accuracy, the $\pm$ shows  $95\%$ confidence intervals over $1000$ test-time  tasks. For computation time, the $\pm$ shows  standard deviation over $1000$ meta iterations.
  %The best performance is highlighted in bold for each meta-learning scenario. 
%   Numbers at upper-level are accuracy with its  $95\%$ confidence interval over 1000 test tasks. Numbers at lower-level are train time per upper-iteration averaged over 1000 iterations run on a 2080S GPU.The base units for accuracy and time are $100\%$ and $1$ second respectively. Sign-MAML achieves better accuracy than MAML for all tasks with less computation time. Sign-MAML also outperforms FO-MAML for all tasks expect for 10-way 5-shot. The computation cost of FO-MAML is comparable to that of Sign-MAML.
  }
  }
  \label{sample-table}
  \centering
  \begin{tabular}{llll}
    \toprule
      Scenario &MAML & FO-MAML 
         & Sign-MAML \\
    \midrule
    \multirow{2}{*}{5-way 1-shot} &35.8 $\pm$ 1.4 $\%$ & 32.7 $\pm$ 1.3 $\%$ & {{37.5 $\pm$ 1.4}} $\pmb{\%}$\\
       & 0.058 $\pm$ 0.003  & {0.032 $\pm$ 0.003}  & {0.032 $\pm$ 0.003} \\
      \hline
    \multirow{2}{*}{5-way 5-shot} & 48.8 $\pm$ 0.7 $\%$ & 45.8 $\pm$ 0.8 $\%$ & {49.5 $\pm$ 0.7}  $\pmb{\%}$\\
      & 0.073 $\pm$ 0.008  & {0.048 $\pm$ 0.006}  & {0.049 $\pm$ 0.006} \\
    \hline
    \multirow{2}{*}{10-way 1-shot} & 20.9 $\pm$ 0.8 $\%$ & 21.4 $\pm$ 0.8 $\%$ & {22.5 $\pm$ 0.8} $\pmb{\%}$\\
     & 0.064 $\pm$ 0.004  & {0.039 $\pm$ 0.003}  & 0.039 $\pm$ 0.003 \\
    \hline
    \multirow{2}{*}{10-way 5-shot} & 29.9 $\pm$ 0.4 $\%$ & {30.9 $\pm$ 0.4} $\pmb{\%}$ & 30.5 $\pm$ 0.5 $\%$\\
    & 0.106 $\pm$ 0.016  & {0.067 $\pm$ 0.015}  & {0.067 $\pm$ 0.016 } \\
    \bottomrule
  \end{tabular}
\label{table: fc100}
\end{table}
%\end{wraptable}

\subsection{Results and Discussions}

% We present MiniImageNet 5-way 1-shot and 5-way 5-shot results in Table \ref{table: miniImagenet_table}. Comparing to FO-MAML, Sign-MAML performs comparably well for 1-shot but better for 5-shot. Its performances are also close to those of MAML. Figure \ref{fig: trainloss} plots the training loss of the three methods for a 5-way 5-shot task on MiniImageNet.
In what follows, we first show the results of Sign-MAML, FO-MAML and MAML on {\FSCIFAR} for  different $N$-way and $K$-shot classification tasks. We then provide a detailed comparison between Sign-MAML and FO-MAML, which fall into the first-order optimization category,   given  various choices of $N$ and $K$ on MiniImageNet. Furthermore, we show  the effectiveness of {\SignMAML} when  different   fine-tuning steps are used. 

\paragraph{{\FSCIFAR} results}

Table \ref{table: fc100}   presents the performance of {\SignMAML} versus {\MAML} and {\FOMAML} on {\FSCIFAR} for 5-way 1-shot classification,  5-way 5-shot  classification, 10-way 1-shot classification and 10-way 5-shot classification. 
Compared to MAML, Sign-MAML performs slightly better for all tasks and takes only half    computation time per iteration.
% For example, it outperforms MAML by $1.7\%$ and $1.6\%$ in accuracy\todo{not obvious since CIs of MAML, SignMAML overlap. May be better to say we are comparable with less time} with $0.026$ and $0.024$ seconds\todo{might be better to talk in terms of \% less time} less time per upper-iteration for 5-way 1-shot classification and 10-way 1-shot classification respectively. 
%In terms of the computation time, it takes 0.026 seconds less for Sign-MAML to run 1 upper-iteration than MAML for 5-way 1-shot classification. 
Compared to {\FOMAML}, Sign-MAML 
achieves a remarkable 
%a similar accuracy for 10-way 5-shot classification, but 
  increase of $4.8\%$ and $3.7\%$  in accuracy for 5-way 1-shot classification and 5-way 5-shot classification, respectively. Moreover, it has very similar computation cost as FO-MAML for all tasks. Overall, Sign-MAML is a competitive method in both performance and computation efficiency when compared to FO-MAML and MAML.

%\SL{[I stop here.]}

\begin{figure}[htb]
%\vspace*{-5mm}
\centerline{
\begin{tabular}{cc}
\hspace*{-6mm}
\includegraphics[width=1.2\textwidth,height=!]{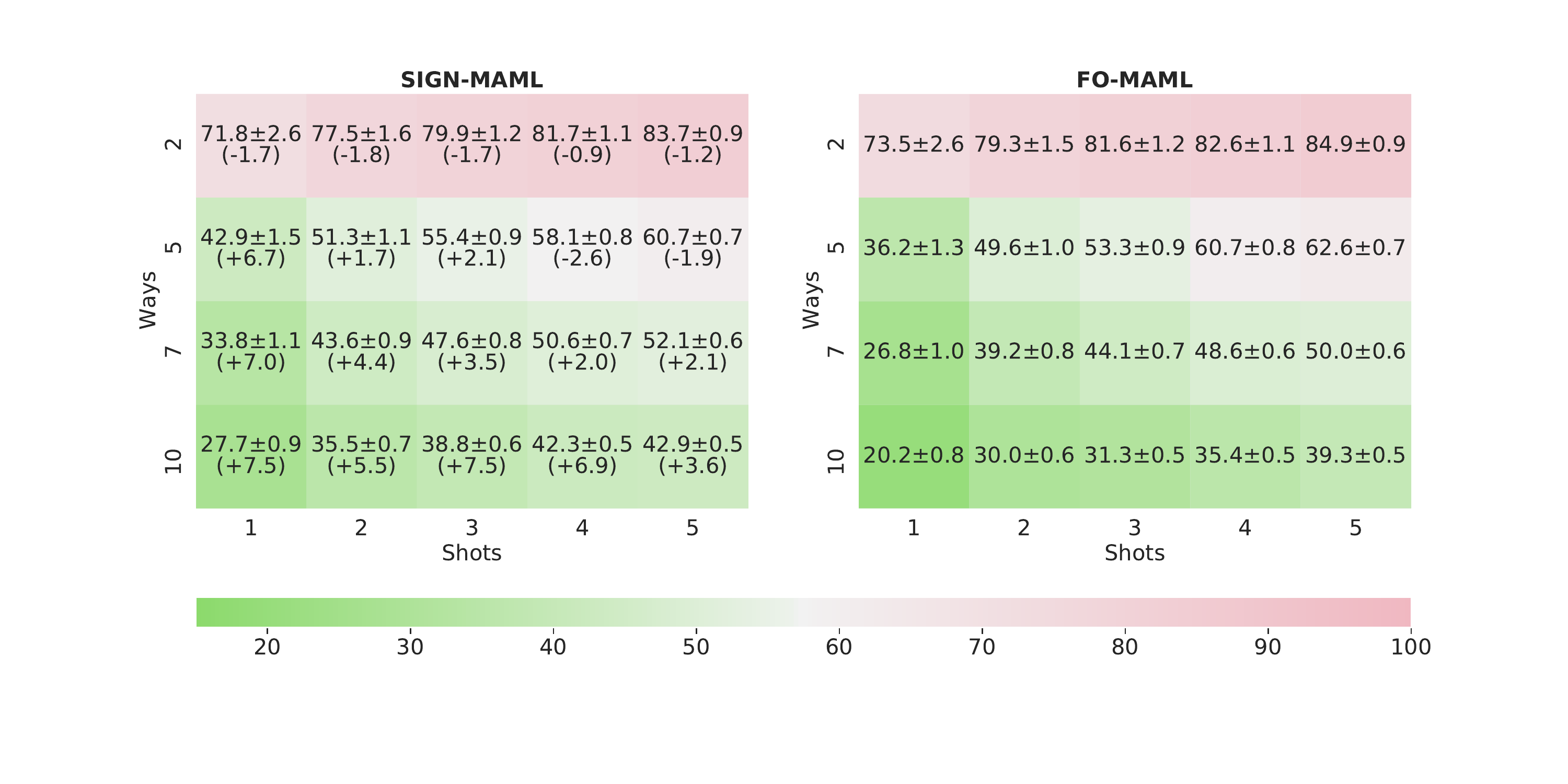} 
%\\ \hspace*{-0.1in}

%\\
%\footnotesize{(a) MiniImageNet 10-way 2-shot} &   \footnotesize{(b) MiniImageNet 5-way 2-shot Time}
\end{tabular}}
\vspace*{-10mm}
\caption{\footnotesize{
MiniImageNet classification results of Sign-MAML and FO-MAML for different ways and shots.
Numbers in each cell are accuracy with its 95 $\%$ confidence interval over test tasks. Numbers inside bracket represent the performance improvement (+) or degradation ($-$) of   Sign-MAML over FO-MAML.  A \textcolor{green}{green} or \textcolor{red}{red}  region indicates the scenario  in which {\SignMAML} is \textcolor{green}{better}  or \textcolor{red}{worsen} than {\FOMAML} in accuracy.
%A positive number means that Sign-MAML outperforms FO-MAML and vice versa.  The base unit is 100 $\%$. 
%Sign-MAML performs better than FO-MAML for tasks of high ways or low shots (i.e. green region). 
}}
\vspace*{-3mm}
\label{fig: heat}
%\vspace{-2em}
%\end{wrapfigure}
\end{figure}

\paragraph{Sign-MAML vs. FO-MAML}
In Figure \ref{fig: heat}, we  present the classification accuracy of {\SignMAML} and {\FOMAML} in a variety of few-shot learning setup, with 
$N \in \{2, 5, 7, 10\}$ ways and $K \in \{1, 2, 3, 4, 5\}$ shots on MiniImageNet. We compare algorithms in the computation-lightest regime using 1 gradient unrolling step. As we can see, if the tasks become more challenging, namely, with
%We observe that for more challenging tasks, i.e. 
higher $N$ and lower $K$, then Sign-MAML performs much better than FO-MAML. For example, Sign-MAML achieves an accuracy that is $7.5\%$ higher than FO-MAML for 10-way 1-shot classification. The performance gap becomes larger as $N$ increases or $K$ decreases.
Moreover, for the cases where FO-MAML outperforms Sign-MAML, the performance gaps (1 - 3\%) are smaller than the cases where Sign-MAML outperforms FO-MAML (1 - 8 \%). The above results suggest that  Sign-MAML can be a better approach when challenging tasks are present. 

\begin{figure}[htb]
%\vspace*{2mm}
\centerline{
\begin{tabular}{cc}
%\vspace*{-5mm}
\hspace*{-6mm}
% \includegraphics[width=.45\textwidth,height=!]{Figures/miniImagenet_7w_2s.pdf} 
% %\\ \hspace*{-0.1in}
% &
\includegraphics[width=.45\textwidth,height=!]{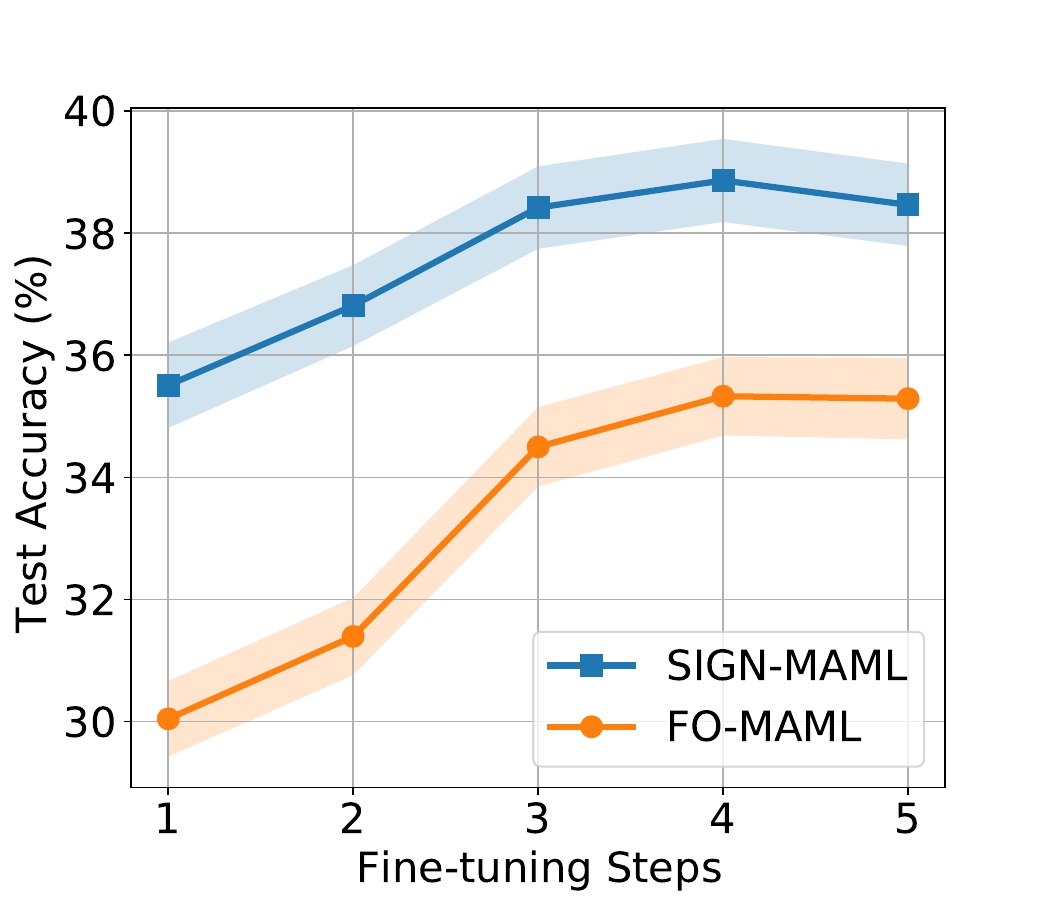}
&  
% \includegraphics[width=.225\textwidth,height=!]{Figures/time_mini_7way_2shot.pdf} 
% &
\includegraphics[width=.45\textwidth,height=!]{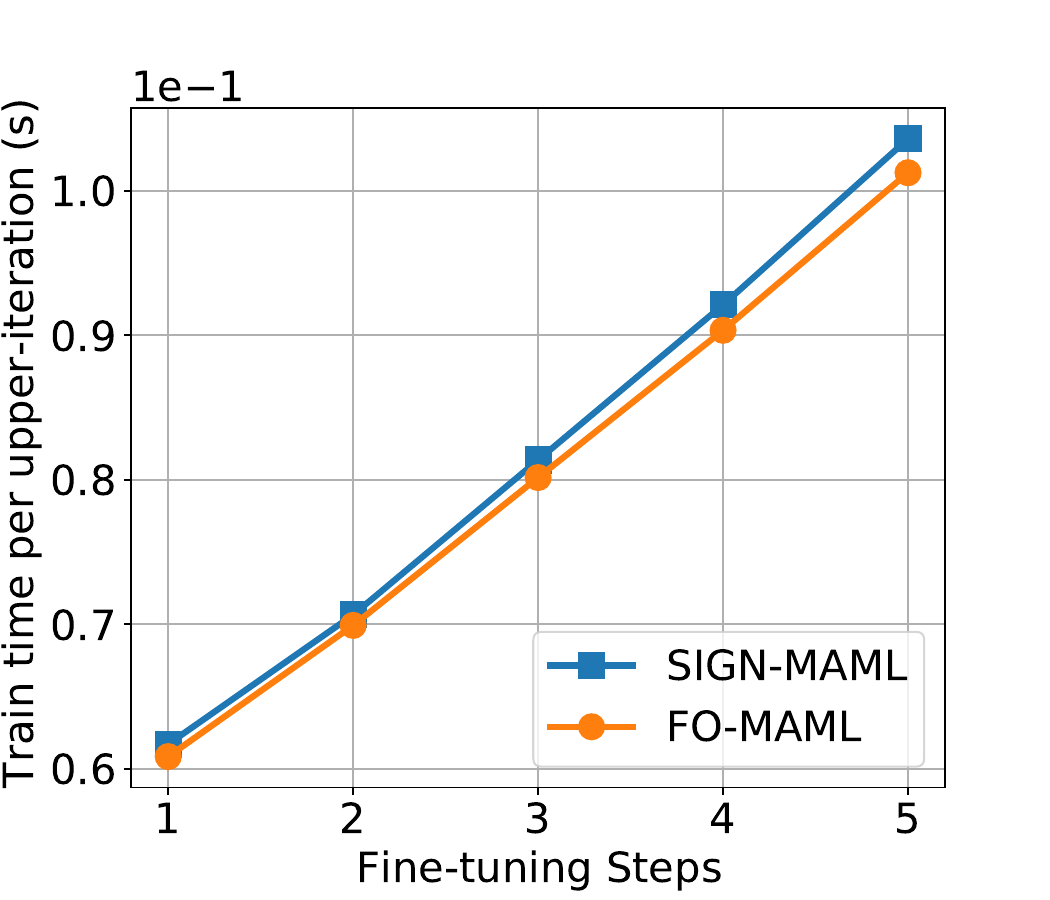} 

\\
% \footnotesize{(a) Accuracy 7-way 2-shot} 
% & 
\footnotesize{(a) Accuracy 10-way 2-shot}
%& \footnotesize{(c) Time 7-way 2-shot}
& \footnotesize{(b) Time 10-way 2-shot}  
\end{tabular}}
%\vspace*{-5mm}

\caption{\footnotesize{
%\SL{Update legends: AT, DAT-PGD, DAT-FGSM}
10-way 2-shot MiniImageNet classification against the choice of the number of fine-tuning steps:
(a) classification accuracy and (b) computation time, with the same format as Table\,1.
% Train time per upper-iteration of these two methods for 7-way 2-shot classification and 10-way 2-shot classification are shown in (c) and (d) respectively. Shaded regions in (a) and (b) show 95 $\%$ confidence interval over test tasks. Time is recorded as the average over 1000 iterations run on a 2080S GPU. Sign-MAML outperforms FO-MAML at each fine-tuning step with similar computation cost.  
}}
\vspace*{1mm}
\label{fig: mini1}
%\vspace{-2em}
%\end{wrapfigure}
\end{figure}

\paragraph{Meta-learning vs. fine-tuning steps} In Figure \ref{fig: mini1}, we  present the classification accuracy as well as the computation cost versus the number of fine-tuning steps. Here we focus on the case of 
%7-way 2-shot classification and 
10-way 2-shot classification on MiniImageNet. 
It can be seen from Figure \ref{fig: mini1} (a)  that
%We observe that 
Sign-MAML outperforms FO-MAML at each setup of the fine-tuning step, and the test accuracy increases rapidly at the beginning and saturates towards the end. 
In addition,
Figure \ref{fig: mini1} (b)  shows that the accuracy improvement of {\SignMAML} over {\FOMAML} is \textit{not} at the cost of computation complexity.
%the train time per upper-iteration for 7-way 2-shot and 10-way 2-shot classification respectively. We observe that 
Clearly,  Sign-MAML and FO-MAML take the very similar computation cost, which  increases linearly with the number of fine-tuning steps (see Appendix C for results of 5-way 2-shot classification and 7-way 2-shot classification on MiniImageNet).
%(See Appendix \todo{If possible, provide precise links to sections, figures, tables in the appendix}for additional results of 5-way 2-shot classification and numerical values of the data in Figure \ref{fig: mini1}).

 %This is expected as Sign-MAML only involves an additional sign operation at lower-level gradient unrolling.

% \begin{figure}[htb]
% %\vspace*{2mm}
% \centerline{
% \begin{tabular}{cccc}
% %\vspace*{-5mm}
% \hspace*{-6mm}
% \includegraphics[width=.225\textwidth,height=!]{Figures/miniImagenet_7w_2s.pdf} 
% %\\ \hspace*{-0.1in}
% &
% \includegraphics[width=.225\textwidth,height=!]{Figures/miniImagenet_10w_2s.pdf}
% &  
% \includegraphics[width=.225\textwidth,height=!]{Figures/time_mini_7way_2shot.pdf} 
% &
% \includegraphics[width=.225\textwidth,height=!]{Figures/time_mini_10way_2shot.pdf} 

% \\
% \footnotesize{(a) Accuracy 7-way 2-shot} & \footnotesize{(b) Accuracy 10-way 2-shot} & \footnotesize{(c) Time 7-way 2-shot} & \footnotesize{(d) Time 10-way 2-shot}  
% \end{tabular}}
% %\vspace*{-5mm}

% \caption{\footnotesize{
% %\SL{Update legends: AT, DAT-PGD, DAT-FGSM}
% MiniImageNet classification results of Sign-MAML and FO-MAML for 7-way 2-shot (a) and 10-way 2-shot (b) at different fine-tuning steps. Train time per upper-iteration of these two methods for 7-way 2-shot classification and 10-way 2-shot classification are shown in (c) and (d) respectively. Shaded regions in (a) and (b) show 95 $\%$ confidence interval over test tasks. Time is recorded as the average over 1000 iterations run on a 2080S GPU. Sign-MAML outperforms FO-MAML at each fine-tuning step with similar computation cost.  
% }}
% \vspace*{1mm}
% \label{fig: mini1}
% %\vspace{-2em}
% %\end{wrapfigure}
% \end{figure}

\paragraph{Take-away:}
Based on the aforementioned results, we find that \textit{\ding{172} Sign-MAML typically performs better than FO-MAML for challenging tasks without losing computation efficiency.  \ding{173}
Sign-MAML can match or exceed the performance of MAML with less computation time.}
%\clearpage

\section{Conclusion}
In this paper, we show that signSGD can be used as an efficient gradient unrolling scheme to advance {\MAML} (model-agnostic meta-learning). Specifically, the study of {\MAML} through the lens of BLO (bilevel optimization) enables us to 
customize  a `lower-level' optimizer to `fine-tune' meta model over task-specific losses. 
We theoretically show that if signSGD is used as the lower-level optimizer, then {\MAML} can be equivalently transformed into  the \textit{first-order} alternating optimization method, termed {\SignMAML}. 
Empirically, we also demonstrate that compared to  the conventional
{\MAML} and  {\FOMAML} approaches,
{\SignMAML} places a more graceful tradeoff between accuracy and computation cost. Particular, in a series of challenging few-shot image classification tasks (which involve more classes and less data samples), {\SignMAML}  yields a consistent improvement over baselines.

% lower-level signSGD unrolling used in bilevel optimization leads to a first-order alternating optimization algorithm. Based on this, we designed a new method for MAML to reduce its computation cost by replacing SGD lower-level unrolling with signSGD lower-level unrolling. We demonstrated that our method achieves Hessian-free meta-gradients naturally without relying on the assumptions made by FO-MAML. We conducted experiments on MiniImageNet and Fewshot-CIFAR100 datasets for different $N$-way $K$-shot classifications to compare the performance of Sign-MAML against FO-MAML and MAML. The experiment results revealed that Sign-MAML can perform better than FO-MAML and MAML with comparable or lower computation cost. Hence, Sign-MAML is an efficient method with high performances. 
% \todo{Might be worthwhile explicitly highlighting that it is interesting/important to understand why Sign-MAML does better than FO-MAML in small $K$ large $N$ setup. The pattern is very clear but we dont have a clear understanding and that this would be part of future work. Unless, this is just for MiniImageNet and not a general pattern.}

% \paragraph{Broader Impact} \todo{Not sure we need Broader Impact for workshop submission}
% Future work involves analyzing the convergence rate of Sign-MAML, testing its empirical performance on large neural networks for few-shot image classification, and exploring its applications in other areas such as natural language processing and reinforcement learning. 

{{
\bibliographystyle{unsrtnat}
\bibliography{refs}
}}

\newpage
\clearpage
%\section{Appendix}
\appendix

 \section{Hyperparameter Search} 
 In this section, we provide more details on hyperparameter tuning for the lower-level learning rate. For Sign-MAML, the search range is $[0.0035, 0.005,0.0065,0.0075,0.01]$; for FO-MAML, the search range is $[0.06,0.08,0.1,0.12,0.14,0.16]$; for MAML, the search range is $[0.06,0.08,0.1,0.12,0.14,0.16]$. If the initial optimal learning rate happens at the end of the range, we continue search in that direction until we find a better one that is within range. For example, if the  optimal learning rate initially found is $0.16$, then we may search $0.18$ and $0.2$. If $0.18$ outperforms $0.16$ and $0.2$, we stop at this point; if $0.2$ outperforms the other two, we repeat the process and search further. 

\section{Train Loss}
\begin{figure}[htb]
%\vspace*{-5mm}
\centerline{
\begin{tabular}{cc}
\hspace*{-6mm}
\includegraphics[width=0.5\textwidth,height=!]{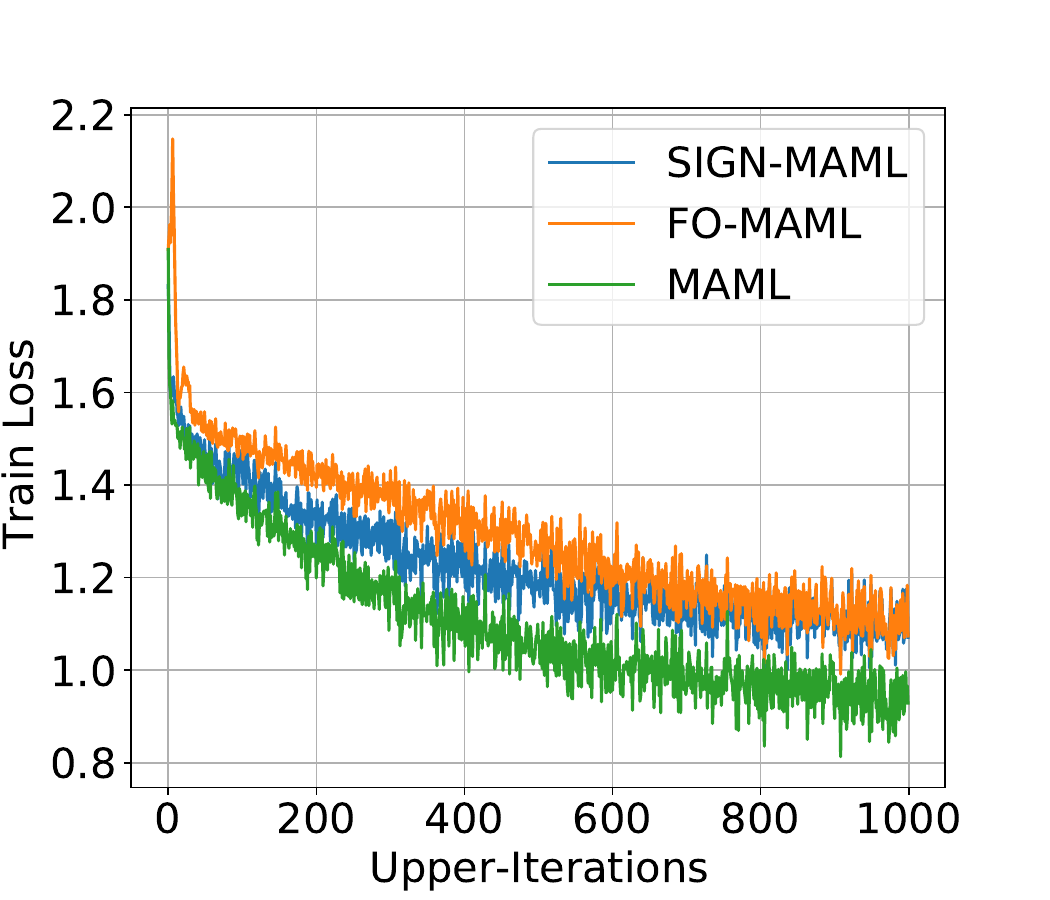} 
%\\ \hspace*{-0.1in}

%\\
%\footnotesize{(a) MiniImageNet 10-way 2-shot} &   \footnotesize{(b) MiniImageNet 5-way 2-shot Time}
\end{tabular}}
%\vspace*{-10mm}
\caption{\footnotesize{
Train loss for MiniImageNet 5-way 5-shot classification. The lower-level learning rates for Sign-MAML, FO-MAML and MAML are 0.005, 0.06 and 0.06 respectively. Meta-batch size is 32.
}}
\vspace*{2mm}
\label{fig: trainloss}
%\vspace{-2em}
%\end{wrapfigure}
\end{figure}

\section{Additional results on MiniImageNet for different fine-tuning steps}

\begin{figure}[htb]
\vspace*{-5mm}
\centerline{
\begin{tabular}{cccc}
%\vspace*{-5mm}
\hspace*{-6mm}
\includegraphics[width=.225\textwidth,height=!]{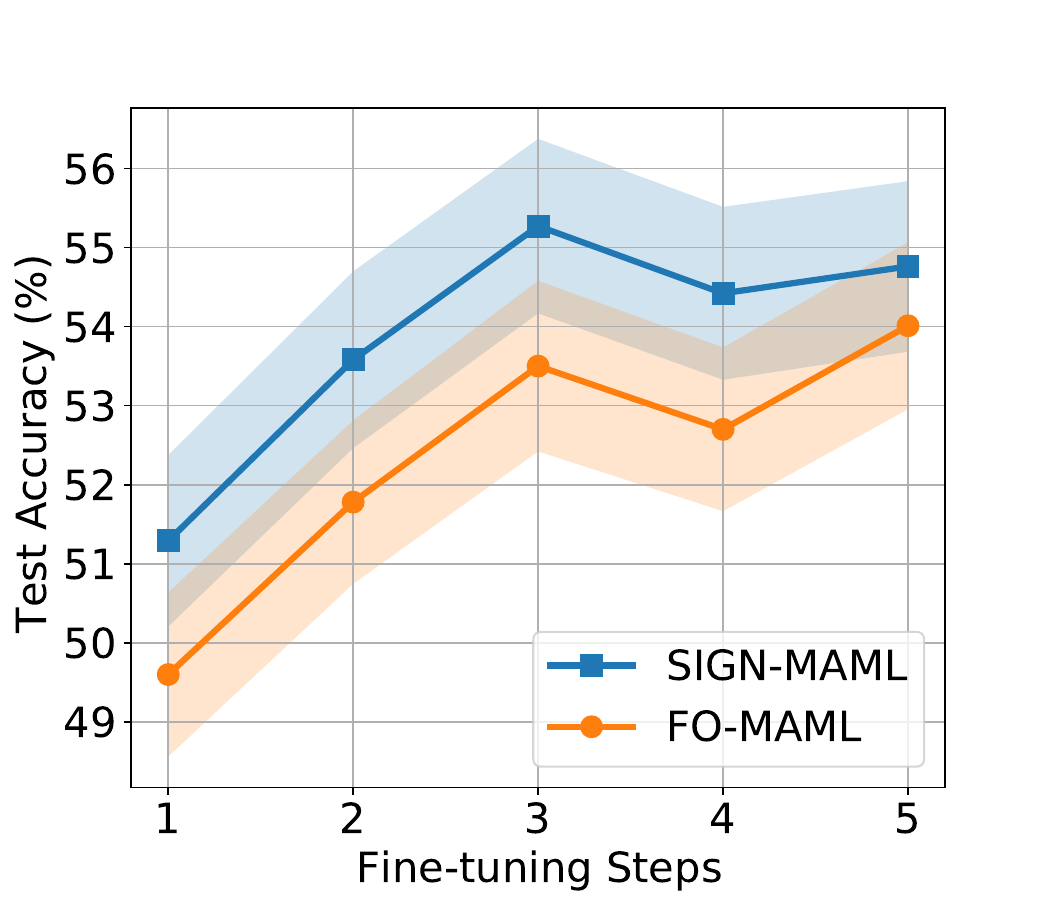} 
%\\ \hspace*{-0.1in}
&
\includegraphics[width=.225\textwidth,height=!]{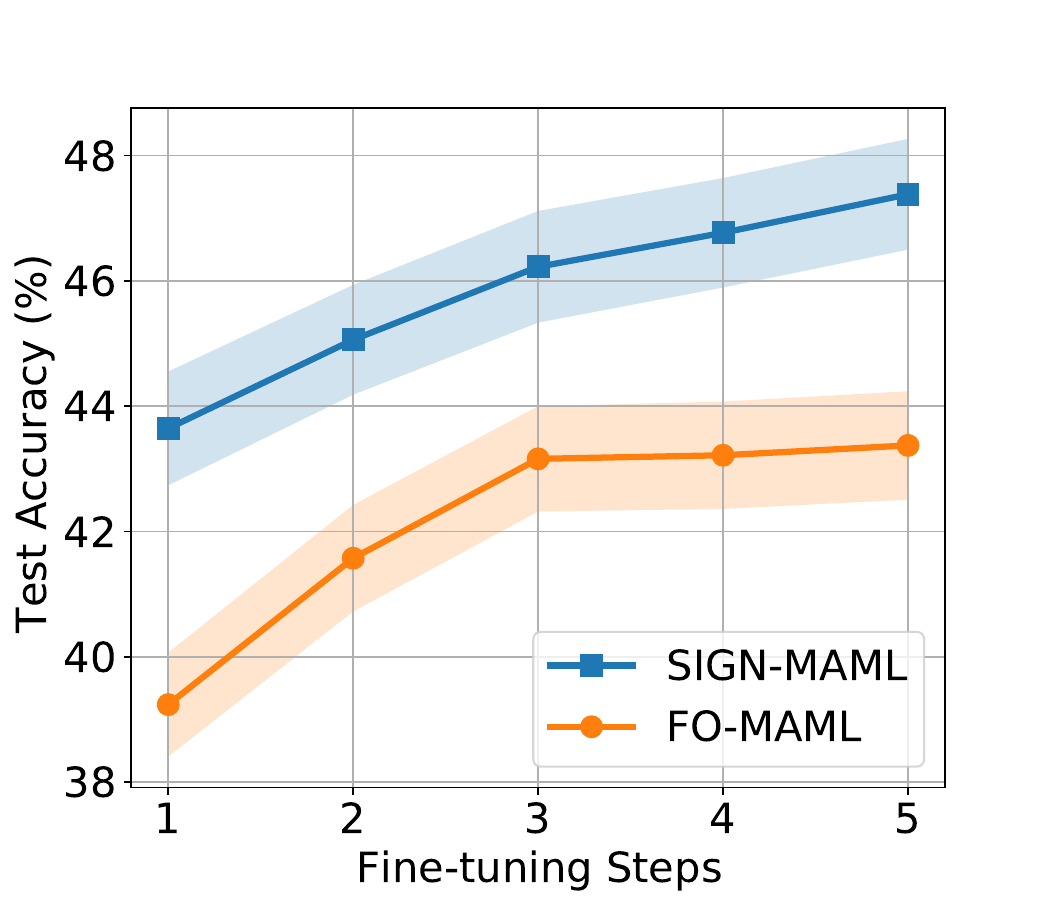}
&
\includegraphics[width=.225\textwidth,height=!]{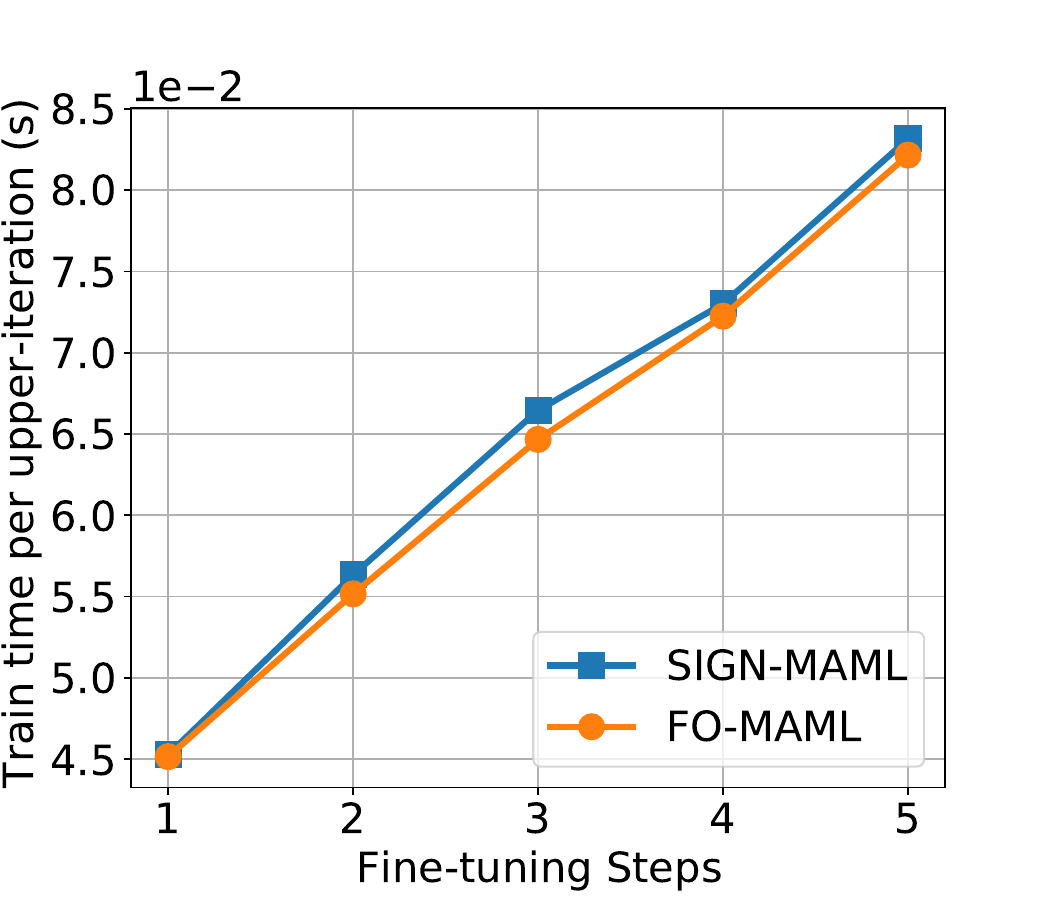} 
%\\ \hspace*{-0.1in}
&
\includegraphics[width=.225\textwidth,height=!]{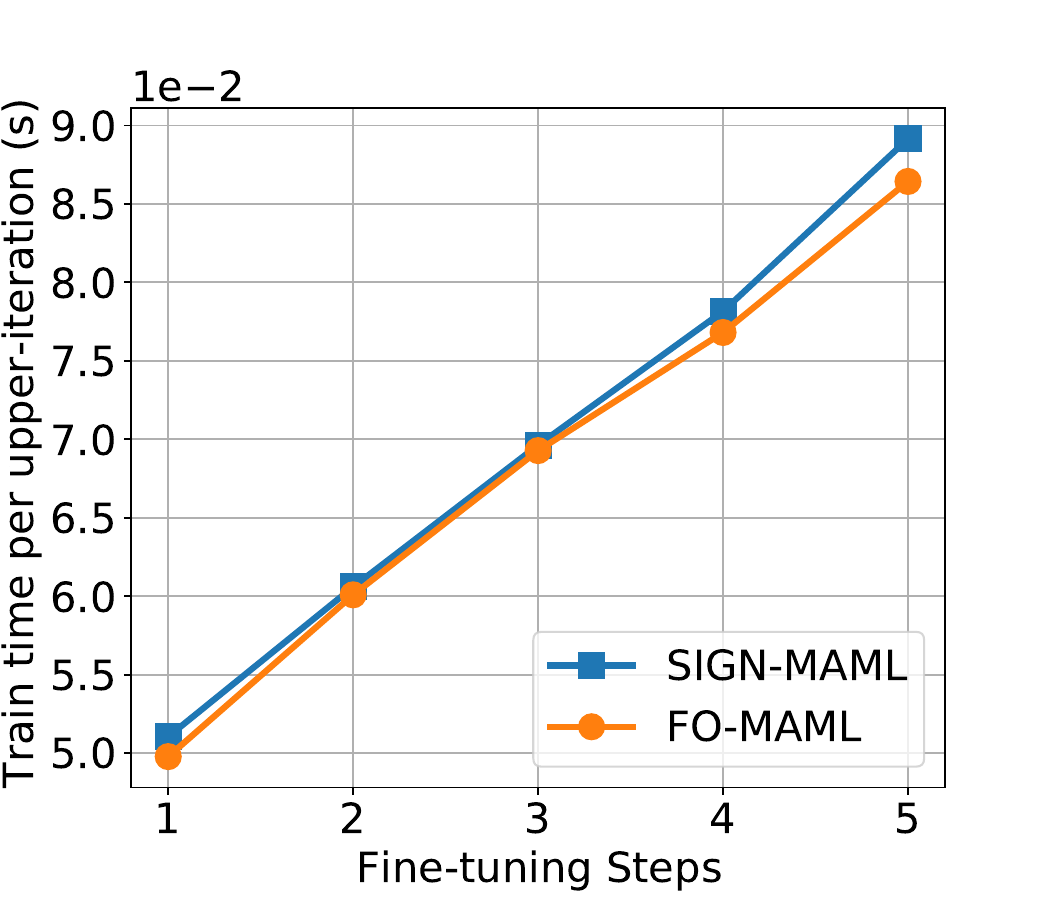}

\\
\footnotesize{(a) Accuracy 5-way 2-shot} &   \footnotesize{(b) Accuracy 7-way 2-shot} & 
\footnotesize{(c) Time 5-way 2-shot} & 
\footnotesize{(d) Time 7-way 2-shot}
\end{tabular}}
%\vspace*{-5mm}

\caption{\footnotesize{
%\SL{Update legends: AT, DAT-PGD, DAT-FGSM}
MiniImageNet classification against the choice of the number of fine-tuning steps:
(a) 5-way 2-shot classification accuracy, (b) 7-way 2-shot classification accuracy, (c) 5-way 2-shot computation time and (d) 7-way 2-shot computation time.
}}
\vspace*{1mm}
\label{fig: mini2}
%\vspace{-2em}
%\end{wrapfigure}
\end{figure}

\end{document}